\documentclass[runningheads]{llncs}
\usepackage[T1]{fontenc}
\usepackage{graphicx,verbatim}
\usepackage{amsfonts}
\usepackage{amsmath} 
\usepackage{algorithm}
\begin{document}
\title{Disentangling Prompt Dependence to Evaluate Segmentation Reliability in Gynecological MRI}
\titlerunning{Disentangled Prompt Dependence}
%
\author{Elodie Germani\inst{1}\orcidID{0000-0002-5786-9538} \and
Krystel Nyangoh-Timoh\inst{1,2} \and
Pierre Jannin\inst{1} \and 
John S.H. Baxter\inst{1}}
\authorrunning{Germani et al.}
%
\institute{Laboratoire Traitement du Signal et de l’Image (LTSI, INSERM UMR 1099), Université de Rennes, Rennes, France \and
Centre Hospitalier Universitaire de Rennes (CHU Rennes), Rennes, France\\
\email{elodie.germani@univ-rennes.fr}}


  
\maketitle              
\begin{abstract}
Promptable segmentation models (\textit{e.g.}, the Segment Anything Models) enable generalizable, zero-shot segmentation across diverse domains. Although predictions are deterministic for a fixed image–prompt pair, the robustness of these models to variations in user prompts, referred to as \textit{prompt dependence}, remains underexplored. In safety-critical workflows with substantial inter-user variability, interpretable and informative frameworks are needed to evaluate prompt dependence. In this work, we assess the reliability of promptable segmentation by analyzing and measuring its sensitivity to prompt variability. We introduce the first formulation of prompt dependence that explicitly disentangles prompt ambiguity (inter-user variability) from local sensitivity (interaction imprecision), offering an interpretable view of segmentation robustness. Experiments on two female pelvic MRI datasets for uterus and bladder segmentation reveal a strong negative correlation between both metrics and segmentation performance, highlighting the value of our framework for assessing robustness. The two metrics have low mutual correlation, supporting the disentangled design of our formulation, and provide meaningful indicators of prompt-related failure modes.


\end{abstract}
\section{Introduction}
Foundation Models~\cite{moor_foundation_2023} have recently reshaped the field of semantic segmentation by separating visual understanding from user input, enabling a single model to generalize across datasets and tasks. The Segment Anything Model (SAM)~\cite{kirillov_segment_2023} exemplifies this paradigm: given an image and a lightweight prompt (\textit{e.g.}, points or bounding boxes), the model outputs a segmentation mask without task- or domain-specific fine-tuning. In medical imaging, this paradigm is of high interest given the scarcity of annotations and the heterogeneity of acquisition conditions. Recent work, such as MedSAM~\cite{ma_segment_2024}, shows that effectively deploying SAM-based models in clinical settings still requires addressing important challenges.

One challenge is that promptable segmentation models' outputs can be highly dependent on the prompt: small changes in a bounding box can lead to large changes in the predicted mask~\cite{rahman_pp-sam_2024}. This effect is exacerbated in medical imaging, where low contrast, artifacts, and diffuse boundaries can induce inter-observer variability even among experts~\cite{andreadis_interobserver_2026,zheng_interobserver_2025}. Moreover, inter-observer variability varies across image modalities and structures: organs with sharp boundaries (\textit{e.g.}, bladder) show higher agreement than organs with smooth or infiltrative boundaries (\textit{e.g.}, uterus)~\cite{zuo_semi-supervised_2026}. In this work, we aim to improve the reliability of promptable segmentation models by analyzing their dependence on prompt variability.

Prompt dependence can be decomposed into two distinct components reflecting different sources of variability. First, \textit{prompt ambiguity} refers to situations where multiple prompts are clinically plausible for a given image and target. This image property captures inter-observer variability in prompt definition due to diffuse boundaries or subjective interpretation. Second, \textit{measurement noise} reflects imprecision: when an observer intends a specific prompt, discretization and drawing variability introduce small perturbations. This property is also intrinsic to the model: even when the target is unambiguous, a model may yield different outputs for minor variations in the prompt.

Previous studies on prompt dependence in segmentation models typically address these components together. To mitigate prompt dependence and improve robustness, prompt refinement methods~\cite{chen_aop-sam_2025,zheng_curriculum_2024} use detection networks to infer an optimal prompt for a given image and label. Similarly, \cite{zhou_medsam-u_2025} proposes iteratively generating prompts and selecting those that maximize a heuristic or an uncertainty-guided criterion, which can improve results but at the expense of increased computational cost. On the contrary, \cite{kaiser_uncertainsam_2025} treats the prompt as fixed to compute uncertainty at the mask level via ensembles or stochastic inference. However, both approaches conflate local sensitivity with prompt uncertainty, leading to prompt refinement strategies that collapse multiple plausible hypotheses into a single prompt or favor consistent yet incorrect solutions, and mask-level uncertainties that appear overconfident by ignoring prompt ambiguity.

Our \textbf{contributions} are threefold: (i) we introduce the first framework for prompt dependence that disentangles prompt ambiguity (inter-user variability) from local sensitivity (interaction imprecision), providing an interpretable perspective on segmentation reliability ; (ii) we model the image-conditioned prompt distribution using a Mixture Density Network (MDN), allowing us to quantify image-level prompt ambiguity and propagate it to pixel-level uncertainty ; (iii) we propose a novel stability margin formulation to estimate local sensitivity under measurement noise, using realistic prompt perturbations to predict the minimal perturbation level required to induce mask changes. We perform an extensive evaluation of two promptable segmentation models in female pelvic MRI, assessing the potential of our framework for reliability and robustness estimation. 

\section{Methods}
\begin{figure}[bt!]
    \centering
    \includegraphics[width=0.9\linewidth]{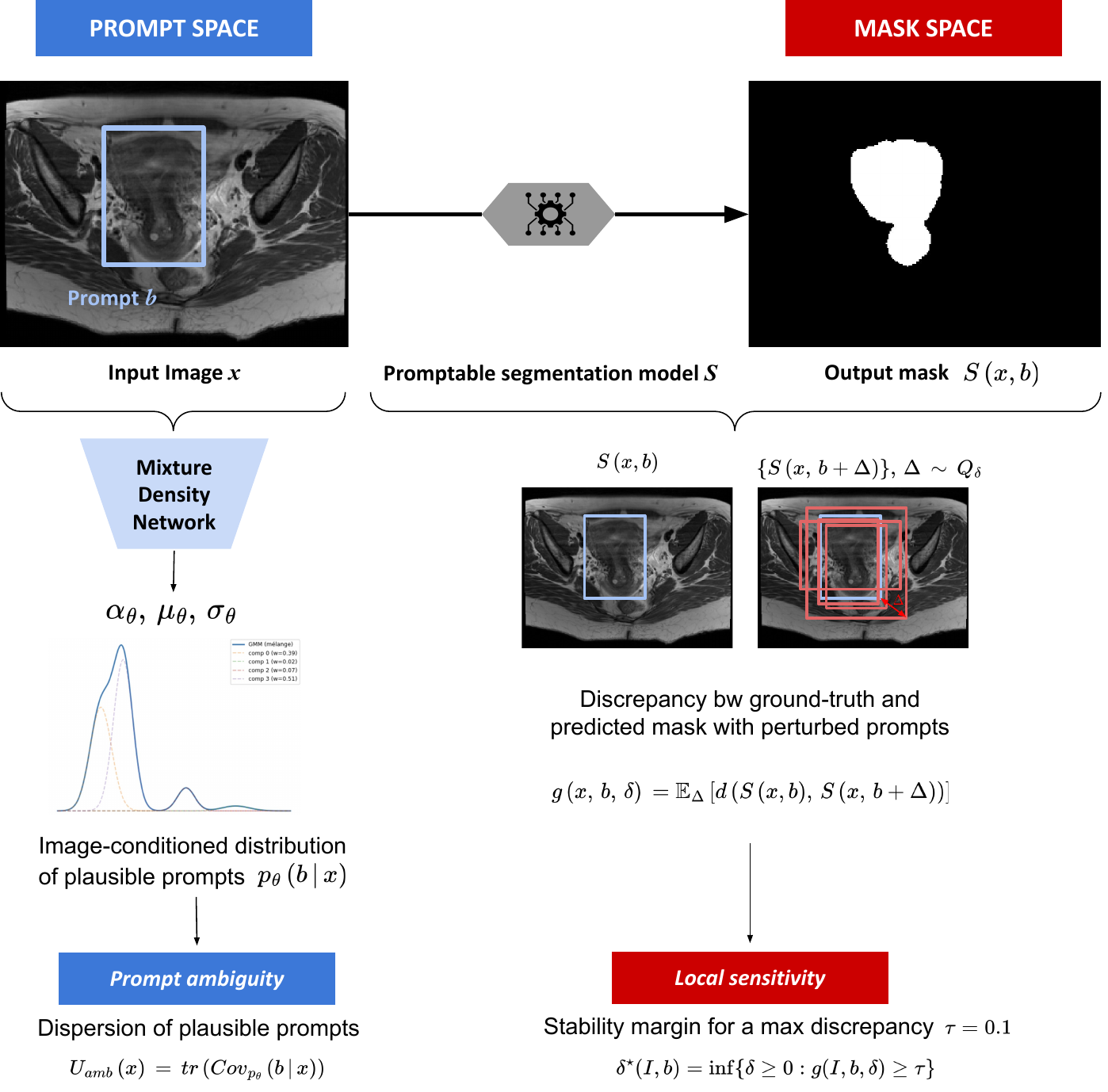}
    \caption{\textbf{Overview of the prompt dependence evaluation framework.} In prompt space (\textit{left}), a conditional mixture model estimates the distribution of plausible prompts $P_{\text{valid}}(b\mid I)$. The dispersion of this distribution, quantified by the trace of its covariance, defines \textbf{prompt ambiguity}. By adding controlled perturbations around reference prompts and measuring their impact in mask space, we train a stability-margin predictor to estimate the minimum change needed to produce a meaningful discrepancy between masks, thereby providing a formulation of \textbf{local sensitivity}. }
    \label{fig:placeholder}
\end{figure}

\subsection{Theoretical framework}
Let $\mathcal{S}: \mathcal{I} \times \mathcal{B} \rightarrow [0,1]^{H \times W}$ 
be a frozen promptable segmentation model mapping an image $I \in \mathcal{I}$ 
and a prompt $b \in \mathcal{B} \subset \mathbb{R}^4$ to a probabilistic mask 
$\hat y = \mathcal{S}(I,b)$.  
Here, both the image $I$ and the model $\mathcal{S}$ are fixed, and we analyze variability in probabilistic masks $\hat {y} $ for a segmentation task $t$ with respect to a distribution of prompts $B$. We explore two levels of variability: between masks generated from different plausible prompts, and local variability around a reference prompt. \\ \\
\textit{\textbf{Variance decomposition.}} For a given image $I$ and task $t$, we assume a distribution of clinically plausible prompts \(B \sim P_{\text{valid}}(b \mid I,t)\) capturing geometric ambiguity. The prompt-marginalized prediction \(\hat Y=\mathcal{S}(I,B)\) is then a random variable conditioned on \((I,t)\), whose variability is quantified by \[\mathrm{Var}(\hat Y \mid I,t) = 
\mathrm{Var}_{B \sim P_{\text{valid}}(\cdot \mid I,t)}
\big[\mathcal{S}(I,B)\big]
\]
To characterize local prompt sensitivity, we further introduce a perturbation variable \(\Delta \sim Q_\delta(\cdot \mid B)\) representing small deviations around a reference prompt \(B\), and define the perturbed prediction \(\hat Y'=\mathcal{S}(I,B+\Delta)\). \\
Applying the law of total variance to \(\hat Y'\) with conditioning variable (B) yields
\[
\mathrm{Var}(\hat Y' \mid I,t) = \mathbb{E}_{B}\left[\mathrm{Var}_{\Delta}\big(\mathcal{S}(I,B+\Delta)\mid B\big)\right]
+
\mathrm{Var}_{B}\left(\mathbb{E}_{\Delta}\big[\mathcal{S}(I,B+\Delta)\mid B\big]\right)
\]
The first term 
\(\mathrm{Var}_{\Delta}\big(\mathcal{S}(I,B+\Delta) \mid B\big)\)  quantifies the variability of the model $\mathcal{S}$ under prompt perturbations $\Delta$ around a reference prompt \(B\). The second term \(
\mathrm{Var}_{B}\left(\mathbb{E}_{\Delta}[\mathcal{S}(I,B+\Delta) \mid B]\right)\), can be approximated for sufficiently small perturbations $\Delta$ by 
\(\mathbb{E}_{\Delta}[\mathcal{S}(I,B+\Delta)\mid B]\approx\mathcal{S}(I,B).\)
Overall, this leads to the following decomposition of the variance of prompt-marginalized predictions:
\[
\mathrm{Var}(\hat Y')
\approx
\underbrace{
\mathbb{E}_{B}
\big[
\mathrm{Var}_{\Delta}
\mathcal{S}(I,B+\Delta)
\big]
}_{\textbf{Local sensitivity}}
+
\underbrace{
\mathrm{Var}_{B}
\big[
\mathcal{S}(I,B)
\big]
}_{\textbf{Prompt ambiguity}}.
\]
By introducing a local perturbation variable and applying the law of total variance, we decompose the variance of prompt-marginalized predictions into (i) the expected local variance induced by prompt perturbations (\textit{local sensitivity}) and (ii) the variance across reference prompts (\textit{prompt ambiguity}). \\ \\
\textit{\textbf{Interpretation of local sensitivity.}} Rather than estimating local sensitivity via infinitesimal perturbations, we define it in terms of a stability margin. For an image $I$ and a reference prompt $b$, we set $d(\cdot,\cdot)$ to be a mask discrepancy measure. For perturbations $\Delta \sim Q_\delta(\cdot \mid b)$ of amplitude $\delta$, we define
\[g(I,b,\delta) = 
\mathbb{E}_{\Delta}
\big[
d(\mathcal{S}(I,b), \mathcal{S}(I,b+\Delta))
\big].\]
Given a tolerance threshold $\tau$ representing the maximum acceptable discrepancy between masks, we define the stability margin as \(\delta^\star(I,b)=\inf\{\delta \ge 0 : g(I,b,\delta) \ge \tau\}.\) This quantity represents the smallest perturbation level required to induce a non-acceptable change in the predicted mask. Larger values of $\delta^\star$ indicate robust segmentation with respect to prompt perturbations. \\ \\
\textit{\textbf{Interpretation of prompt ambiguity.}} 
For a fixed image $I$, let $B \sim P_{\text{valid}}(b \mid I)$ denote a random plausible prompt and define $f(b)=\mathcal{S}(I,b)$. The prompt-marginalized prediction $f(B)$ is then a random variable with variability \(\mathrm{Var}_B[f(B)].\)
Assuming $f$ is locally differentiable in a neighborhood of $\mu_B=\mathbb{E}[B]$ and that $B$ has covariance $\Sigma_B$, a first-order expansion gives
\[
f(B) \approx f(\mu_B) + \nabla_b f(\mu_B)^\top (B-\mu_B)
\Rightarrow
\mathrm{Var}_B[f(B)] \approx \nabla_b f(\mu_B)^\top \Sigma_B \nabla_b f(\mu_B).
\]
This approximation makes explicit that prompt-induced variability increases with the dispersion of the valid-prompt distribution in prompt space (captured by $\Sigma_B$) and with the local prompt sensitivity (captured by $\nabla_b f(\mu_B)$).

\subsection{Empirical estimation}
\textbf{\textit{Prompt ambiguity.}} In practice, the true distribution of valid prompts $P_{\text{valid}}(b \mid I)$ is unknown. We approximate it with a conditional density model \(p_\theta(b \mid x), \, x = \mathrm{pool}(E(I)),\) where $E$ is the frozen image encoder and $p_\theta$ is implemented as a mixture density network (MDN)~\cite{bishop_mixture_1994,mclachlan_mixture_1988} (see Fig.~\ref{fig:placeholder}, left). 
The MDN is trained by conditional maximum likelihood on pairs $(x_i, b_{i})$. This parametric approximation of the image-conditioned prompt distribution enables estimation of prompt ambiguity via the predictive covariance of the learned mixture, which captures both intra-mode dispersion and inter-mode separation \(
U_{\text{amb}}(I)=\mathrm{tr}\big(\mathrm{Cov}_{p_\theta}(b \mid x)\big)\). \\ \\ 
\textbf{\textit{Local sensitivity.}} We estimate the stability margin $\delta^\star(I,b)$ for a given image around a representative plausible prompt $b$ sampled from $p_\theta(b \mid x)$. We evaluate $g(I,b,\delta)$ over a discrete set of perturbation levels and select the smallest $\delta$ such that the average mask difference exceeds $\tau=0.1$. This empirical procedure requires no additional ground-truth masks and relies solely on the model's response to synthetic, clinically relevant perturbations. To enable efficient inference, we use a pretrained stability-margin predictor $r_\psi$, that provides a fast approximation of local sensitivity without iterative sampling (see Fig.~\ref{fig:placeholder}, right): \(\delta^\star(I,b) = r_\psi(I,b,\tau) \, \text{with} \, \tau=0.1\). \\ \\
The proposed framework provides computable estimators of the two theoretical components of prompt-induced variability:
(i) ambiguity of valid prompts, modeled explicitly via a conditional mixture density, and 
(ii) local sensitivity under measurement noise, captured by the stability margin. 

\section{Experiments and Results}

\paragraph{\textbf{Datasets}.} We evaluate the proposed framework on two datasets of T2-weighted pelvic MRI acquired in the axial plane: UT-EndoMRI~\cite{liang_multi-modal_2025} and MOGaMBO~\cite{manna_federated_2025}. UT-EndoMRI contains $N=81$ female patients with endometriosis and manual uterus segmentations, while MOGaMBO includes $N=94$ female patients with locally advanced cervical cancer and manual bladder segmentations. We apply a standard preprocessing pipeline: 3D volumes and labels converted into 2D axial slices, resampled to a spacing of $1.0\times1.0$\, mm, and reoriented to a common coordinate space. Pixel intensities are min-max normalized per slice, and only slices containing organ annotations are retained. Each selected slice is resized to $1024\times1024$ and exported as a 3-channel image. We derive a ground-truth bounding box $b={x_1,y_1,x_2,y_2}$ from each manual mask and reparameterize it for stable learning as $u={\mathrm{logit}(x_c),\mathrm{logit}(y_c),\log H,\log W}$, where $(x_c,y_c)$ are normalized box-center coordinates and $(H,W)$ its height and width. Datasets are split at the patient level into 70\%/10\%/20\% train/val/test, with respectively $\{773,501,255\}$ and $\{195,128,62\}$ slices for MOGaMBO and UT-EndoMRI.

\paragraph{\textbf{Implementation}.} We apply our framework for prompt dependence assessment on two frozen SAM-based architectures: MobileSAM~\cite{zhang_faster_2023} and MedSAM~\cite{ma_segment_2024}. Image embeddings $x$ are extracted for each 2D slice using each image encoder (with $dim(x)=512$), and ground-truth bounding boxes are computed from manual segmentation masks. For prompt ambiguity, we train a conditional mixture density network (MDN) implemented as a two-layer multi-layer perceptron (MLP) with ReLU activations, taking as input the mean-max pooled image embedding and predicting the parameters of a Gaussian mixture distribution $p_\theta(b \mid x)$ with $K=8$ components, including mixture weights, means, and diagonal covariances. The number of components has been chosen by fitting standard GMM with various $K$ values and the value giving the lowest BIC score on the validation set was chosen. The MDN is trained with a negative log-likelihood (nll) loss, the Adam optimizer with a learning rate of $1\times10^{-3}$, and a batch size of 8 for 100 epochs. The parameters giving the lowest nll on the validation sets were selected. The stability margin predictor is also implemented as a two-layer MLP that takes as input the concatenation of the image embedding, bounding box, and the tolerance threshold $\tau$. Oracle stability margins are computed offline using 32 samples per perturbation type (translation or scale) and level over a discrete grid of $\delta$ values ($\delta \in \{0.005, 0.01, 0.05\}$). The model learns to predict stability margins $\delta^\star$ using an $\ell_1$ loss, an Adam optimizer with a learning rate of $1\times10^{-3}$, and a batch size of 64 for 100 epochs. All models are implemented in PyTorch v2.4.1. 

\begin{figure}[t!]
    \centering
    \scriptsize
    UT-EndoMRI \hspace{4cm} MOGaMBO \\ 
    \includegraphics[width=0.9\linewidth]{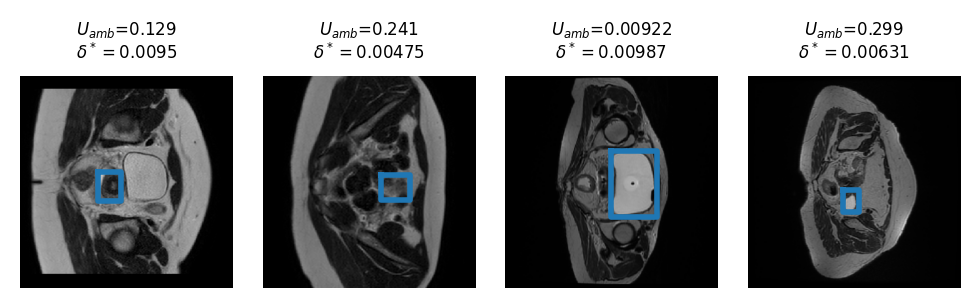} \\ 
    \resizebox{\textwidth}{!}{
    \begin{tabular}{c|c|c|c|c|c|c}
    \hline
    {Dataset} & {$Dice$ {\tiny (SEM)}} & { $U_{amb}$ {\tiny(min, max)}} & { $\delta^\star$ {\tiny(min, max)}} & { $r$ \tiny $(U_{amb}, \delta^\star)$} & { $r$ \tiny $(U_{amb}, Dice)$} & { $r$ \tiny ($\delta^\star$, $Dice$)}\\ 
    \hline
    & \multicolumn{6}{c}{{MobileSAM~\cite{zhang_faster_2023}}} \\ 
    \hline
    {\tiny MOGaMBO} & 0.81 \textit{\tiny (0.004)} & 0.14 \textit{\tiny (0.02-0.34)} & 0.01 \textit{\tiny (0.003-0.05)} & -0.07 & -0.45$^*$ & 0.36$^*$ \\
    \hline
    UT-EndoMRI & 0.86 \textit{\tiny (0.004)} & 0.19 \textit{\tiny (0.12-0.23)} & 0.01 \textit{\tiny (0.003-0.02)} & -0.001 & -0.30$^*$ & 0.44$^*$ \\
    \hline
    & \multicolumn{6}{c}{{MedSAM~\cite{ma_segment_2024}}} \\ 
    \hline
    {\tiny MOGaMBO} & 0.33 \textit{\tiny (0.07)} & 0.07 \textit{\tiny(0.002-0.37)} & 0.003 \textit{\tiny(0.0002-0.02)} & -0.04 & -0.30$^*$ & 0.21$^*$ \\
    \hline
    {\tiny UT-EndoMRI} & 0.80 \textit{\tiny (0.005)} & 0.21 \textit{\tiny (0.18-0.27)} & 0.0001 \textit{\tiny ($1e^{-5}$-0.0004)} & 0.05 & -0.37$^*$ & -0.46$^*$ \\
    \hline
    \end{tabular}}
    \caption{Qualitative examples (top) and dataset-level summary statistics (bottom) of prompt dependence metrics, $U_{amb}$ and $\delta^\star$, for UT-EndoMRI and MOGaMBO. For each dataset, two representative slices are shown with the corresponding $U_{amb}$ and $\delta^\star$ values; the blue boxes highlight the ground-truth bounding box. The table reports the mean Dice (±SEM), the ranges of $U_{amb}$ and $\delta^\star$, and Pearson correlations on the test set. Stars $\star$ indicate significant correlations at $p<0.001$.}\label{fig:metrics}
\end{figure}

\subsection{Prompt ambiguity and local sensitivity in organ segmentation}
Both the conditional MDN and the margin predictor were trained successfully, achieving $\text{nll}\approx -2.5$ and $\ell_1\approx 0.001$ on validation sets.
Fig.~\ref{fig:metrics} (top panel) displays representative samples from each dataset, their corresponding ground-truth bounding boxes, and prompt dependency metrics from MobileSAM experiments. Bounding box locations and sizes vary across samples and organs, but do not appear to be associated with ambiguity or sensitivity. Organ shapes and boundaries, however, appear irregular and blurry in samples with high $U_{amb}$. Regarding local sensitivity, samples with low stability margins $\delta^\star$ tend to have smaller bounding boxes that are closely adjacent to adjacent structures, which could explain the low perturbation level required to induce mask changes at $\tau=0.1$. Similar observations were made for experiments with MedSAM. Average metrics values for each dataset and model are shown in Tab.~\ref{fig:metrics} (bottom panel). Across both datasets and models, prompt ambiguity and local sensitivity exhibit low correlations ($r_{MOG} =-0.07$ and $r_{UT} = 0.001$ with MobileSAM), supporting the complementary and distinct aspects of prompt dependence. 

\subsection{Prompt dependence as an indicator of segmentation quality}
\begin{figure}[htb!]
    \centering
    \scriptsize
    MobileSAM \\ 
    UT-EndoMRI \hspace{3cm} MOGaMBO \\ 
    \includegraphics[width=0.43\linewidth]{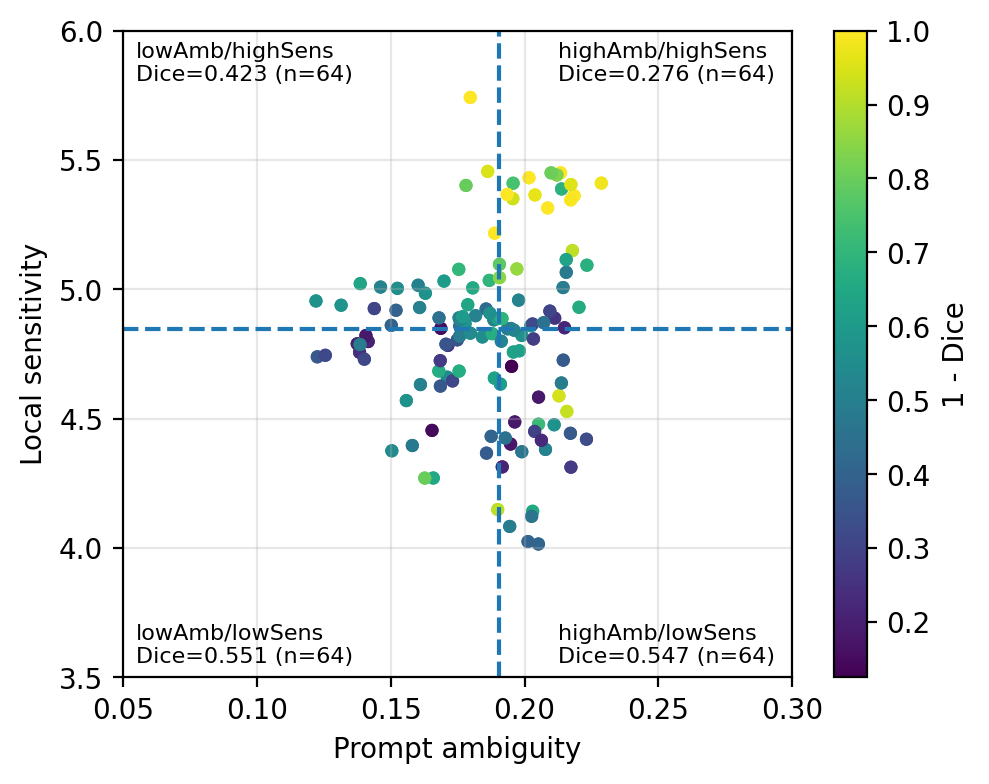} \includegraphics[width=0.43\linewidth]{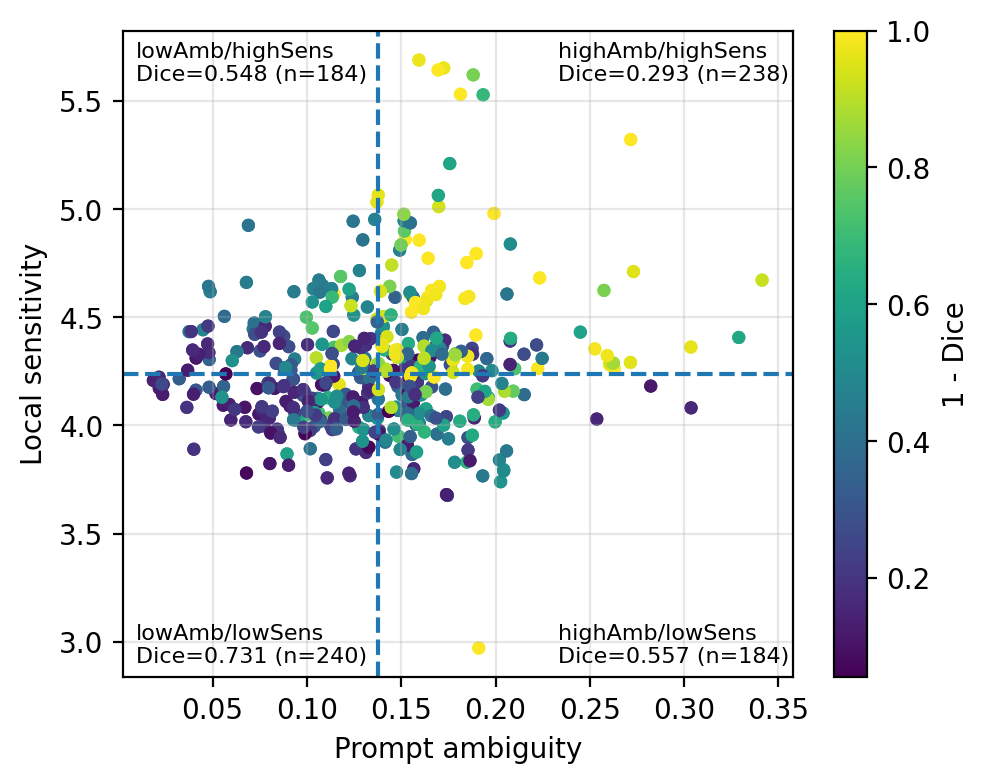} \\ 
    \includegraphics[width=0.9\textwidth]{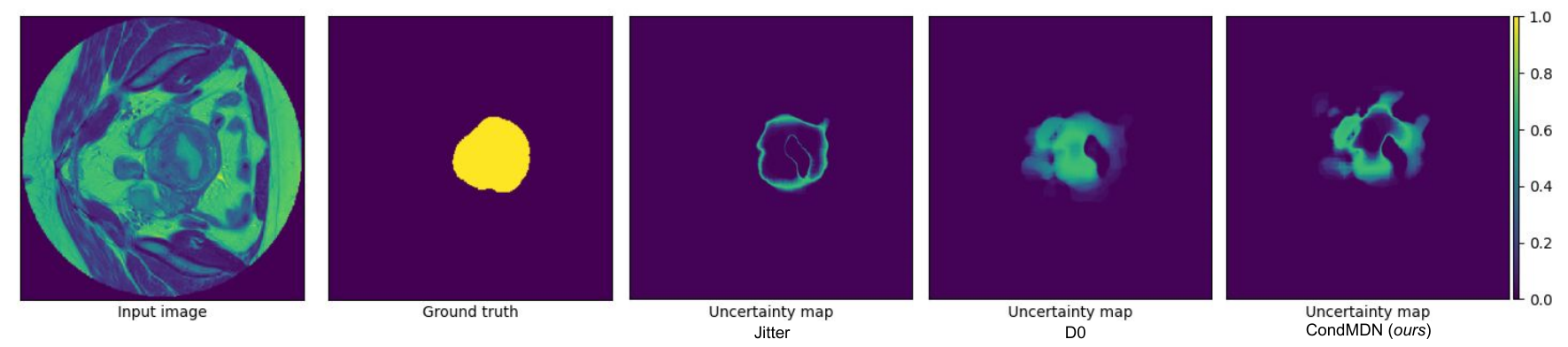} \\ 
    \begin{tabular}{c|c|c|c|c}
         Method & AUROC $\, \uparrow$ & Entropy$_{mean}$ & $\Delta\text{Entropy}_\text{ok/err}\, \uparrow$ & $nll \, \downarrow$ \\
         \hline
        D0 & 0.96 & 0.05 & \textbf{0.49} & 0.09 \\
        Jitter & 0.93 & 0.03 & 0.35 & 0.12 \\
        CondMDN (ours) & \textbf{0.98} & 0.07 & 0.45 & \textbf{0.05} \\
    \end{tabular}
    \caption{(\textit{Top}) Distribution of prompt ambiguity (x-axis) and local sensitivity (y-axis, encoded as $\log\big(\frac{1}{1-\delta^\star}\big)$), color-coded by $1-Dice$. Dashed lines denote the median values used to define the four classes. (\textit{Bottom}) Qualitative example of uncertainty maps produced by Jitter, DO, and CondMDN with the input image and ground truth mask on UT-EndoMRI, with averaged evaluation metrics of uncertainty maps.}
    \label{fig:dice}
\end{figure}

We use the median values of $U_{amb}$ and $\delta^\star$ within each dataset (Fig.~\ref{fig:dice}, top panel) to define four classes of prompt dependence. Overall, lower segmentation quality is achieved in images with both high prompt ambiguity and high local sensitivity ($+/+$) than the other classes ($\overline{Dice}_{+/+}=0.29$, $\overline{Dice}_{-/-}=0.73$ for MOGaMBO with MobileSAM). The association between prompt dependence metrics and segmentation quality is further supported by the inverse correlations of each metric with Dice scores (\textit{e.g.}, $r({Dice,U_{amb})}=-0.45$ for MOGaMBO and $r{(Dice,U_{amb})}=-0.30$ for UT-EndoMRI with MobileSAM). However, the level of correlation varies across datasets, models, and metrics: in MOGaMBO, Dice scores are more strongly anti-correlated with $U_{amb}$ than with $\delta^\star$ ($r=0.36$). Conversely, in UT-EndoMRI, Dice scores are more strongly anti-correlated with $\delta^\star$ ($r=-0.44$) than with $U_{amb}$ ($r=-0.30$). Similar, though less pronounced, results are observed in experiments with MedSAM (see Fig.~\ref{fig:dice}, top panel). These results suggest that the main driver of prompt dependence is dataset-specific and differently encoded by image encoders, reflecting complementary failure modes and differences in image quality or organ appearance.

\subsection{From prompt ambiguity to pixel-level uncertainty}
Beyond prompt-ambiguity scores, \(p_\theta(b\mid x)\) enables a wide range of analyses. Here, we propose using this distribution to derive pixel-wise uncertainty maps that represent uncertainty due to inter-user variability. For each image, we sample \(N=1000\) prompts, decode the corresponding masks, and compute the per-pixel entropy across the resulting mask ensemble. Fig.~\ref{fig:dice} (bottom) illustrates uncertainty maps obtained with three strategies: (i) jittering a fixed, user-provided prompt, (ii) sampling prompts from a non-conditional distribution \(D_0\), and (iii) sampling from the conditional model \(p_\theta(b\mid x)\). In contrast to standard perturbation-based approaches that primarily capture local prompt sensitivity, conditional sampling marginalizes over the space of clinically plausible prompts, yielding spatially coherent uncertainty concentrated along ambiguous anatomical boundaries. Quantitatively, uncertainty maps derived from \(p_\theta(b\mid x)\) show high spatial calibration, with high entropy consistently aligned with regions of segmentation errors (on average, $\Delta\text{Entropy}_\text{ok/err}=0.45$ in UT-EndoMRI, and \(0.38\) in MOGaMBO for uncertainty maps with MobileSAM).

\section{Conclusion}

In this work, we propose to evaluate the reliability of medical image segmentation by focusing on prompt variability, characterized by inter-user variability in prompt definition, increased in ambiguous cases, and by measurement imprecision due to local sensitivity in image-model settings. Our metrics correlate significantly with segmentation quality and enable interpretable evaluation of failure modes in promptable segmentation. Further experiments are needed to better understand the impact of hyperparameters, starting with the number of components used to model $p_\theta(b \mid x)$ and the stability margin threshold $\tau$. Next steps include discussions on how best to integrate this evaluation framework into clinical practice, to improve target delineation in radiotherapy, and to extract robust organ and lesion biomarkers from images. Our code is publicly available at: \url{https://anonymous.4open.science/r/PromptDependence-3039}.
\begin{credits}

\end{credits}

%
%
%
%

\end{document}